\documentclass[11pt]{article}
\usepackage{nlprs01}
\usepackage{epsf}
\title{Part of Speech Tagging in Thai Language \\ Using Support Vector Machine}
\author{Masaki Murata, Qing Ma, and Hitoshi Isahara\\
Communications Research Laboratory \\
{\normalsize 2-2-2 Hikaridai, Seika-cho, Soraku-gun, Kyoto, 619-0289, Japan}\\
{\normalsize \{murata,qma,isahara\}@crl.go.jp}}

\begin{document}
\maketitle
\begin{abstract}
The elastic-input neuro tagger and hybrid tagger, combined
with a neural network and Brill's error-driven learning, have already been proposed 
for the purpose of constructing a practical tagger using as little training 
data as possible. 
When a small Thai corpus is used for training, 
these taggers have tagging accuracies of 94.4\% and 95.5\% 
(accounting only for the ambiguous words in terms of the part of speech), respectively. 
In this study, in order to construct more accurate taggers 
we developed new tagging methods 
using three machine learning methods: the decision-list, maximum entropy, and support vector machine methods. 
We then performed tagging experiments by using these methods. 
Our results showed that 
the support vector machine method has the best precision (96.1\%), 
and that it is capable of improving the accuracy of tagging in the Thai language. 
Finally, we theoretically examined all these methods and 
discussed how the improvements were achived. 
\end{abstract}

\section{Introduction}

The elastic-input neuro tagger and hybrid tagger, combined 
with a neural network and Brill's error-driven learning, have already been proposed 
for the purpose of constructing a practical tagger using as little training 
data as possible. 
When a small Thai corpus is used for training, 
these taggers have tagging accuracies of 94.4\% and 95.5\% 
(accounting only for the ambiguous words in terms of the part of speech), respectively. 
In this study, in order to construct more accurate taggers we developed new tagging methods 
using three machine learning methods: the decision-list, maximum entropy, and support vector machine methods. 
We then performed tagging experiments by using these methods. 
As supervised data for POS tagging in the Thai language we used 
the same corpus as in our group's previous papers \cite{ma_col98,ma_ijcnn99,ma_coling2000}. 

In connection with our approach, we should emphasize the following points:
\begin{itemize}
\item 
  In this work, we perfomed POS tagging in the Thai language 
  by using the support vector machine method. 
  Although many studies have considered 
  POS tagging by using machine learning methods, 
  few studies have used the support vector machine method. 
  This method achieves high perfomance, 
  but it requires huge machine resources and does not work 
  when we use large-scale corpora as supervised data. 
  In addition, 
  with large-scale corpora 
  we can obtain good performance by using a simple method 
  such as HMM (hidden Markov model). 
  For the Thai language, however, 
  large-scale corpora have not yet been constructed, 
  so our apporach is effective. 

\item 
  We also carried out experiments by using the decision list 
  and maximum entropy methods for comparison, 
  and we confirmed that 
  the support vector machine method produced the best precision. 
  This paper shows data comparing the performace. 

\item 
  The precision produced by the support vector machine method 
  was slightly higher than that obtained in a previous study \cite{ma_coling2000}, 
  which used the hybrid tagger combined 
  with a neural network and Brill's error-driven learning. 
  Since our precision was slightly higher, 
  we have improved the technology of POS tagging in the Thai language. 
\end{itemize}

\section{Problems with POS tagging}
\label{sec:mondai_settei}

This study did not consider the segmentation of 
a sentence into words. 
We assumed that the words had been segmented before 
POS tagging began.\footnote{The Thai language is 
an agglutinative language like Japanese, 
and it thus has the problem of word segmentation 
in addition to POS tagging in morphological analysis. 
This study did not consider word segmentation. 
To handle word segmentation, 
we have to make all possible segmentations by using a word dictionary 
and then perform a Viterbi search so that 
the probability for POS tagging and word segmentation 
in the whole sentence is as high as possible. 
This study focused on POS tagging, which 
would be one component of the Viterbi search. 
Because our approach uses machine learning methods, 
the probabilities were output with estimated results. 
Thus we can easily use this study as one component in the Viterbi search.}
In this case, a sentence is expressed as follows:
\begin{equation}
S = (w^1, w^2, \cdot\cdot\cdot, w^n), 
\end{equation}
where $w^i$ is the $i$-th word in the sentence. 
POS tagging is the application of a POS tag to each word. 
Therefore, the result of POS tagging is expressed as follows:
\begin{equation}
T = (t^1, t^2, \cdot\cdot\cdot, t^n)
\end{equation}
where $t^i$ is the tag for the POS of word $w^i$. 
Our goal is to determine the correct POS tag for each word. 
The categories indicated by the POS tags are defined in advance. 
POS-tagging problems can thus be regarded as classification problems 
and can be handled by machine learning methods. 

\section{Machine learning methods}
\label{sec:ml_method}

In this paper, 
we used the following three machine learning methods:\footnote{Although 
there are also such decision-tree learning methods as C4.5, 
we did not use them for the following two reasons. 
First, decision-tree learning methods perform worse than 
the other methods 
on several tasks \cite{murata_coling2000,taira_svm_eng}. 
Second, the number of attributes used in this research 
was very large, and 
the performance of C4.5 would become worse if 
the number of attributes was decreased 
so that C4.5 could work.}
\begin{itemize}
\item 
  decision-list method
\item 
  maximum-entropy method
\item 
  support-vector machine method
\end{itemize}
In this section, 
these machine-learning methods are explained. 

\subsection{Decision-list Method}

In this method, 
pairs consisting of a feature $f_j$ and a category $a$ are stored 
in a list, called {\it a decision list}. 
The order in the list is defined in a certain way, and 
all the pairs are arranged in this order. 
The decision list method searches for pairs 
from the top of the list 
and outputs the category of the first pair 
with the same feature as a given problem as the desired answer. 
In this study, we use the value of $p(a|f_{j})$ to 
arrange pairs in order. 

This decision list method is equivalent to 
the following method using probabilistic equations. 
The probability of each category is calculated 
by using one feature $f_j (\in F, 1\leq j\leq k)$, and 
the category with the highest probability is judged to be 
the correct category. 
The probability of 
producing a category $a$ in a context $b$ is given 
by the following equation: 
{
\begin{eqnarray}
  \label{eq:decision_list}
  p(a|b) = p(a|f_{max}),
\end{eqnarray}
}
where $f_{max}$ is defined as
{
\begin{eqnarray}
  \label{eq:decision_list2}
  f_{max} = argmax_{f_j\in F} \ max_{a_i\in A} \ \tilde{p}(a_i|f_j),
\end{eqnarray}
}
such that $\tilde{p}(a_i|f_j)$ is the occurrence rate of 
category $a_i$ when the context includes feature $f_j$. 

\subsection{Maximum-entropy Method}

In this method, 
the distribution of probabilities $p(a,b)$ 
when equation (\ref{eq:constraint}) is satisfied and 
equation (\ref{eq:entropy}) is maximized 
is calculated. 
The category with the maximum probability 
as calculated from this 
distribution of probabilities is judged to be 
the correct category \cite{ristad97,ristad98}: 

{\small
\begin{eqnarray}
  \label{eq:constraint}
  \sum_{a\in A,b\in B}p(a,b)g_{j}(a,b) 
  \ = \sum_{a\in A,b\in B}\tilde{p}(a,b)g_{j}(a,b)\\
  \ for\ \forall f_{j}\ (1\leq j \leq k) \nonumber
\end{eqnarray}
}
{\small
\begin{eqnarray}
  \label{eq:entropy}
  H(p) & = & -\sum_{a\in A,b\in B}p(a,b)\ log\left(p(a,b)\right),
\end{eqnarray}}
where $A, B,$ and $F$ are a set of categories, a set of contexts, 
and a set of features $f_j (\in F, 1\leq j\leq k)$, respectively; 
$g_{j}(a,b)$ is a function with a value of 1 when context $b$ includes feature $f_j$ 
and the category is $a$, and a value of 0 otherwise; 
and $\tilde{p}(a,b)$ is the occurrence rate of 
pair $(a,b)$ in the training data. 

In general, the distribution of $\tilde{p}(a,b)$ is very sparse. 
We cannot use it directly, 
so we must estimate the true distribution of $p(a,b)$ 
from the distribution of $\tilde{p}(a,b)$. 
In the maximum-entropy method, 
we assume that the estimated value of the frequency of 
each pair of category and feature 
calculated from $\tilde{p}(a,b)$ is the same as 
that calculated from $p(a,b)$ (This corresponds to Equation \ref{eq:constraint}.). 
These estimated values are not so sparse. 
We can thus use the above assumption to calculate $p(a,b)$. 
Furthermore, we maximize the entropy 
of the distribution of $\tilde{p}(a,b)$ to 
obtain one solution of $\tilde{p}(a,b)$, 
beacause using only Equation \ref{eq:constraint} produces 
many solutions for $\tilde{p}(a,b)$. 
Maximizing the entropy makes 
the distribution more uniform, which 
is known to provide a strong solution to data sparseness problems. 

\begin{figure}[t]
      \begin{center}
      \epsfile{file=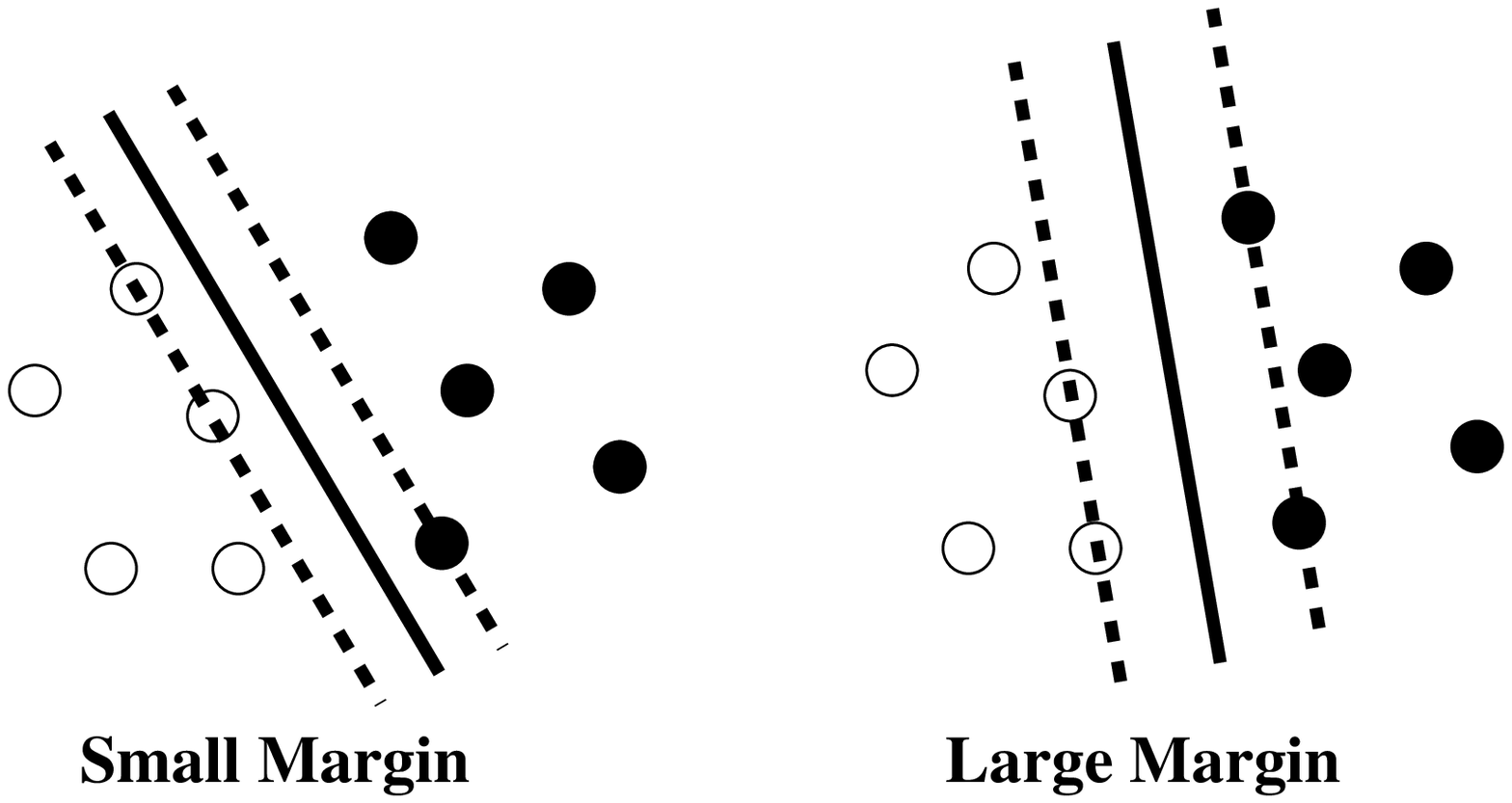,height=4cm,width=8cm} 
      \end{center}
      \caption{Maximizing the margin}
    \label{fig:margin}
\end{figure}

\subsection{Support-vector Machine Method}

In this method, 
data consisting of two categories is classified 
by dividing space with a hyperplane. 
When the two categories are positive and negative and 
the margin between positive 
and negative examples in the training data 
is larger (see Figure \ref{fig:margin}\footnote{In the figure, 
the white circles and black circles indicate 
positive and negative examples, respectively. 
The solid line indicates the hyperplane dividing space, 
and the broken lines indicate planes at 
the boundaries of the margin regions.}), 
the probability of incorrectly choosing 
categories in open data is thought to be smaller. 
The hyperplane maximizing the margin is determined, 
and classification is done by using this hyperplane. 
Although the basics of the method are as described above, 
for extended versions of the method, in general, 
the inner region of the margin in the training data 
can include a small number of examples, and 
the linearity of the hyperplane is changed to non-linearity 
by using kernel functions. 
Classification in the extented methods is equivalent to 
classification using the following discernment function, and 
the two categories can be classified 
on the basis of whether the output value of the function is positive 
or negative \cite{SVM,kudoh_svm}: 

{
\begin{eqnarray}
  \label{eq:svm1}
  f({\bf x}) & = & sgn \left( \sum^{l}_{i=1} \alpha_i y_i K({\bf x}_i,{\bf x}) + b \right)\\
  b & = & -\frac{max_{i,y_i=-1}b_i + min_{i,y_i=1}b_i}{2}\nonumber\\
  b_i & = & \sum^l_{j=1} \alpha_j y_j K({\bf x}_j,{\bf x}_i), \nonumber
\end{eqnarray}
}
where ${\bf x}$ is the context (a set of features) 
of an input example; 
${\bf x}_{i}$ and $y_i (i=1,...,l, y_i\in\{1,-1\})$ indicate 
the context of the training data and its category, respectively; and the function $sgn$ is defined as 
{
\begin{eqnarray}
  \label{eq:svm2}
  sgn(x) \, = & 1 & (x \geq 0),\\
          & -1 & (otherwise). \nonumber
\end{eqnarray}
}
Each $\alpha_i (i=1,2...)$ is fixed 
when the value of $L(\alpha )$ in Equation (\ref{eq:svm4}) 
is maximum under the conditions of 
Equations (\ref{eq:svm5}) and (\ref{eq:svm6}). 
{\small
\begin{eqnarray}
  \label{eq:svm4}
  L({\alpha}) & = & \sum^l_{i=1} \alpha_i - \frac{1}{2} \sum^l_{i,j=1} \alpha_i \alpha_j y_i y_j K({\bf x_i},{\bf x_j})
\end{eqnarray}
}
{
\begin{eqnarray}
  \label{eq:svm5}
  0 \leq \alpha_i \leq C \, \, (i=1,...,l)
\end{eqnarray}
}
{
\begin{eqnarray}
  \label{eq:svm6}
  \sum^l_{i=1} \alpha_i y_i = 0 
\end{eqnarray}
}
Although the function $K$ is called a kernel function and 
various types of kernel functions can be used, 
this paper uses a polynomial function as follows: 
{
\begin{eqnarray}
  \label{eq:svm3}
  K({\bf x},{\bf y}) & = ({\bf x}\cdot{\bf y} + 1)^d,
\end{eqnarray}
}
where $C$ and $d$ are constants set by experimentation. 
In this paper, 
$C$ is fixed as 1 for all experiments. 
Two values of $d$, $d=1$ and $d=2$, are used. 
A set of ${\bf x}_i$ that satisfies $\alpha_i > 0$ is called 
a support vector, and 
the portion used to perform the sum in Equation (\ref{eq:svm1}) 
is calculated by only using examples that are support vectors. 

Support-vector machine methods can handle data 
consisting of two categories. 
In general, 
data consisting of more than two categories can be handled 
by using the pair-wise method \cite{kudoh_chunk_conll2000}. 
In this method, 
for data consisting of N categories, 
all pairs of two different categories (N(N-1)/2 pairs) 
are constructed. 
Better categories are determined 
by using a 2-category classifier (in this paper, 
a support-vector machine\footnote{We use the software 
TinySVM \cite{kudoh_svm} developed by Kudoh as the support-vector machine.}
is used as the 2-category classifier.), and 
finally the correct category is determined 
on the basis of ``voting'' on the N(N-1)/2 
pairs analyzed with the 2-category classifier. 

The support-vector machine method used in this paper is 
in fact implemented by combining the support-vector machine method 
and the pair-wise method described above. 

\section{Features (information used in classification)}
\label{sec:sosei}

Although we have explained the three machine-learning methods, 
using these methods 
requires defining the features 
(information used in classification). 
In this section, we explain these features. 

As mentioned in Section \ref{sec:mondai_settei}, 
when the result of word segmentation of a sentence in Thai language is input, 
we output the POS for each word. 
Therefore, the features are extracted 
from the input Thai sentence. 
Here, we define the following items as features. 
\begin{itemize}
\item 
  {\bf POS information}

  The candidate POS tags of the current word, 
  the three previous words, and the three subsequent words\footnote{\label{fn:kanou_hinshi} In general, 
    since the words preceding the current word have already been analyzed, 
    we can use only the one POS used in the current context, not possible POSs. 
    In fact, previous studies used 
    the POSs of the results of tagging in the previous context. 
    This paper, however, uses possible POSs in the previous context 
    for the following two reasons. One is the easeness of processing, 
    and the other is that 
    we considered cases when the tagging in the previous context 
    was performed wrongly.}(e.g., ``noun'', ``verb'', etc. The total number of 
  features in the Thai corpus is mentioned in Section \ref{sec:experiment}.)

  The candidate POSs were determined in advance for each word by using 
  a word dictionary or the Thai corpus. 
  
\item 
  {\bf POS and order information}
  
  The pair of candidate POS tags and their occurrence order 
  in the current word, 
  three previous words, and three subsequent words\footnote{\label{fn:hinshi_juni} In Ma's previous studies 
    the probability of a POS for each word was used. 
    The machine learning methods (decision list method and maximum entropy method) based on features as used in this paper, 
    however, are difficult to use with continual values such as probabilities in the features. 
    Therefore, we used the occurrence order instead of the occurrence probability. 
    Since the order information is at most the number of ambiguities in POS and 
    thus not so large, 
    the machine learning methods used in this paper can handle the order. 
    On the other hand, 
    the support vector machine methods can handle continual values in the features. 
    However, we used the occurrence order rather than the occurrence probability to enable comparison to 
    the decision list and maximum entropy methods. 
    In the future, we should use the occurrence probability in the support vector machine.}
  (e.g., ``noun, the first place'', ``verb, the second place'', etc.  The total number of such features is 782.)

  The occurrence order indicates the frequency order of the POS 
  in the training data when it is used for the current word. 

\item 
  {\bf word information}

  The current word, 
  three previous words, and three subsequent words
  (e.g., ``tommorow'', ``go'', etc. The total number of such features is 15,763.)

\end{itemize}

\section{Experiments}
\label{sec:experiment}

This section describes 
our experiments on POS tagging in the Thai language 
by using the machine-learning methods described in Section \ref{sec:ml_method} 
with the feature sets described in Section \ref{sec:sosei}, 
for the tasks described in Section \ref{sec:mondai_settei}. 

The experiments in this paper were performed 
by using the same Thai corpus as in our previous papers \cite{ma_col98,ma_ijcnn99,ma_coling2000}. 
This corpus contains 10,452 sentences randomly 
divided into two sets: one with 8,322 sentences, for training; and 
the other with 2,130 sentences, for testing. 
The training and testing sets contain, respectively, 22,311 and 6,717 ambiguous words 
(in other words, 
the target words for POS tagging).\footnote{The total numbers of 
words including non-ambiguous words are 124,331 and 34,544, respectively.}.
The ambiguous words are those that may serve as more than one POS. 
The other words always serve as the same POS, 
and they were assigned to a POS by using a word dictionary 
rather than a machine learning method. 
47 POSs are defined for the Thai corpus \cite{orchid97}. 

\begin{table}[t]
\caption{Experimental results}
\label{tab:result}
  \begin{center}
\renewcommand{\arraystretch}{1.000}
\begin{tabular}[c]{|l|c|}\hline
\multicolumn{1}{|c|}{Method}  & \multicolumn{1}{|c|}{Precision} \\\hline
Baseline method & 83.6\% \\
HMM  & 89.1\% \\
Rule-based       & 93.5\% \\
Elastic NN              & 94.4\% \\
Hybrid tagger      & 95.5\% \\
Decision list           & 83.6\% \\
Maximum entropy    & 95.3\% \\
Support vector machine& 96.1\% \\\hline
\end{tabular}

\end{center}
(Precisions are as obtained for ambiguous words only.)
\end{table}

The experimental results are shown in Table \ref{tab:result}. 
The precisions for ``Baseline method'', ``HMM'', ``Rule-based'', ``Elastic NN'', 
and ``Hybrid tagger'' are from previous papers \cite{ma_ijcnn99,ma_coling2000}. 
In the baseline method, 
a word is judged to represent the POS that 
most frequently appears for that word in the training corpus. 
HMM refers to a method that performs POS tagging at the sentence level 
by using the hidden Markov model. 
``Rule-based'' indicates Brill's method, that is, 
the use of error-driven transformation rules. 
``Elastic NN'' is a method our group proposed previously \cite{ma_ijcnn99}, 
using a three-layered perceptron in which 
the length of the input layer is changeable. 
``Hybrid tagger'' is another method our group proposed previously \cite{ma_coling2000}, 
combining the elastic NN and rule-based methods. 
It improves elastic NN by using Brill's error-driven learning. 
The precision of hybrid tagger was the best 
among our previous studies based on the Thai corpus used in this paper. 
The results in Table \ref{tab:result} for the other three methods (decision list method, maximum entropy method, 
and support vector machine method) were obtained in this study. 

Among these three methods, 
the precision of the support vector machine method (96.1\%\footnote{The precisions shown in this paper 
were obtained using ambiguous words only. 
The precision for all words, including non-ambiguous words, 
was 99.2\%.}) was the best. 
This result is consistent with our other previous studies \cite{murata_acl01_modal,murata_nlc2001_wsd_eng}. 
The precision of the support vector machine method 
was also higher than that of hybrid tagger (95.5\%), which had produced the best precisions 
in the previous studies. 
Therefore 
our study has improved the technology of POS tagging in the Thai language. 

Next, we compared the various methods. 
We first examined the three methods used in this paper. 
Since they used exactly the same features, 
the comparison was strict. 
The order of these methods was as follows:
\begin{quote}
\mbox{\hspace*{0.5cm}$\mbox{Support vector} > \mbox{Maximum entropy}$    }

\hspace*{0.5cm}\hspace*{0.5cm}$> \mbox{Decision list}$    
\end{quote}
The precision of the decision list method was very low and 
almost the same as that of the baseline method. 
This was because we did not use AND features 
(combination of features) as inputs for the system.
We can thus say that by using only one feature the experiments were 
under adverse conditions for the decision list method. 
If we use AND features, 
the precision of the decision list method will increase,\footnote{A previous paper \cite{murata_coling2000} 
showed that the decision list method can produce high precisions for bunsetsu identification 
in Japanese sentences by using AND features. 
In this study, the precision of the decision list method was bad because 
we did not use AND features.}
but when we make AND features randomly, 
the number of features increases explosively. 
When we add a small number of features, 
we need to throughly examine which combinations of features 
must be added. 
In contrast, the support vector and 
maximum entropy methods perform estimation by using all features. 
Furthermore, the support vector machine method 
has a framework for considering AND features automatically 
by adjusting the constant $d$ in the kernel function. 
We can thus say that 
the support vector machine method is an effective machine learning method 
in that we do not have to examine AND features by hand. 

Next, we compared our methods with the previous methods. 
We have to do this carefully, 
because the features used here did not match those used in 
the previous studies. 
We first compared the rule-based and hybrid tagger methods. 
These methods use not only POS information 
but also word information 
in the rule templates used in error-driven learning. 
We can thus say that 
these methods use almost the same features as in this study, 
and therefore, they can be compared to the methods used here. 
We can say that the order of the main machine learning methods 
was as follows:\footnote{Strictly speaking, 
hybrid tagger used the AND features, while 
maximum entropy method can produce better precision when AND features are used. 
Thus, the order of ``Hybrid tagger'' and ``Maximum entropy'' could be changed.}
\begin{quote}
\hspace*{0.5cm}$\mbox{Support vector} > \mbox{Hybrid tagger}$    

\mbox{\hspace*{0.2cm}\hspace*{0.5cm}$> \mbox{Maximum entropy} > \mbox{Rule-based}$    }
\end{quote}

\begin{table}[t]
\caption{Experimental results when word information was eliminated}
\label{tab:result_noword}
  \begin{center}
\renewcommand{\arraystretch}{1.000}
\begin{tabular}[c]{|l|c|}\hline
\multicolumn{1}{|c|}{Method}  & \multicolumn{1}{|c|}{Precision} \\\hline
Decision list            & 78.0\% \\
Maximum entropy   & 92.3\% \\
Support vector machine& 93.9\% \\\hline
\end{tabular}

\end{center}
(Precisions are as obtained for ambiguous words only.)

\end{table}

Next we examined the HMM and elastic NN methods. 
These methods do not use word information directly: 
they only use the probability of the occurrence of a POS in each word. 
We carried out our experiments by eliminating the features of word information 
to create similar conditions for these methods, 
as shown in Table \ref{tab:result_noword}. 
All methods produced lower precision in this case 
than when using word information. 
When we compared elastic NN (94.4\%) and support vector machine (93.9\%) with no word information, 
the former had higher precision. 
Elastic NN, however, uses 
the probability of the occurrence of a POS in each word, 
while support vector machine uses word and order information instead. 
Since this provides less information than 
the probability of the occurrence of a POS, 
this is not a strict comparison. 
However, from these results we expect 
that elastic NN should have performance as high as 
that of support vector machine.\footnote{Although we have compared methods 
using different features, 
we should conduct experiments in which the features are the same.}
As for HMM, we can say that 
it has lower performance than the support vector machine and 
maximum entropy methods, because 
its precision was much lower than for both of these methods.

Finally we examined the reasons why we could 
improve the precision. 
The reason that the support vector machine method produced 
higher precision than the HMM and Elastic NN methods is 
that it uses word information as well. 
(``HMM'' and ``Elastic NN'' did not use word information as mentioned above.)
In some cases 
a POS is determined by a word in the previous or subsequent context, 
and in many of these cases the word information is very helpful. 
Next, we compared the support vector machine method to rule-based and hybrid tagger methods. 
Since almost the same information was used among them, 
we can expect that the support vector machine method should have better performance than the other methods. 
Since hybrid tagger includes Brill's error-driven learning, that is the rule-based method, 
the performance of hybrid tagger will deteriorate 
when the performance of the rule-based method is bad. 
We can thus say that 
we obtained better precision 
because we used word information and 
a support vector machine with good performance. 
As for future work, 
we should conduct experiments by using 
word information in the elastic NN method. 

\section{Conclusions}

In this paper, we examined 
POS tagging in the Thai language 
by using supervised machine learning methods. 
As supervised data we used 
the corpus described in our group's previouse papers \cite{ma_coling2000}.
We used the decision list method, the maximum entropy method, 
and the support vector machine method as machine learning methods. 
In the experimental results, 
the support vector machine method produced the best precision.  
Its precision was 
slightly higher than the precision obtained in a previous study, 
which used a hybrid tagger combined with 
a neural network and Brill's error-driven learning. 

We examined and compared various machine learning methods, including those in previous studies. 
We discussed the good performance of the support vector machine method. 
We expected that 
elastic NN, which is one method from the previous studies, would also 
have good performance, 
but it does not use word information and 
its precision was lower than that of the support vector machine mthod. 
We can say that our method in this paper produced better precision 
because we used word information and 
because we used the support vector machine method whose performance is good. 
For the future work, 
we should conduct experiments by using 
word information in elastic NN method.

%
%

{
\bibliographystyle{acl}
\bibliography{mysubmit}}

\begin{thebibliography}{}

\bibitem[\protect\citename{Charoenporn \bgroup et al.\egroup }1997]{orchid97}
Thatsanee Charoenporn, Virach Sornlertlamvanich, and Hitoshi Isahara.
\newblock 1997.
\newblock Building a large {Thai} text corpus - part-of-speech tagged corpus:
  Orchid -.
\newblock In {\em NLPRS'97}.

\bibitem[\protect\citename{Cristianini and Shawe-Taylor}2000]{SVM}
Nello Cristianini and John Shawe-Taylor.
\newblock 2000.
\newblock {\em An Introduction to Support Vector Machines and Other
  Kernel-based Learning Methods}.
\newblock Cambridge University Press.

\bibitem[\protect\citename{Kudoh and Matsumoto}2000]{kudoh_chunk_conll2000}
Taku Kudoh and Yuji Matsumoto.
\newblock 2000.
\newblock Use of support vector learning for chunk identification.
\newblock {\em CoNLL-2000}.

\bibitem[\protect\citename{Kudoh}2000]{kudoh_svm}
Taku Kudoh.
\newblock 2000.
\newblock {TinySVM: Support Vector Machines}.
\newblock http://cl.aist-nara.ac.jp/~taku-ku// software/TinySVM/ index.html.

\bibitem[\protect\citename{Ma \bgroup et al.\egroup }1998]{ma_col98}
Qing Ma, Kiyotaka Uchimoto, Masaki Murata, and Hitoshi Isahara.
\newblock 1998.
\newblock A multi-neuro tagger using variable lengths of contexts.
\newblock In {\em 17th International Conference on Computational Linguistics
  (COLING-ACL'98)}, pages 802--806.

\bibitem[\protect\citename{Ma \bgroup et al.\egroup }1999]{ma_ijcnn99}
Qing Ma, Kiyotaka Uchimoto, Masaki Murata, and Hitoshi Isahara.
\newblock 1999.
\newblock Elastic neural networks for part of speech tagging.
\newblock In {\em IJCNN'99}.

\bibitem[\protect\citename{Ma \bgroup et al.\egroup }2000]{ma_coling2000}
Qing Ma, Masaki Murata, Kiyotaka Uchimoto, and Hitoshi Isahara.
\newblock 2000.
\newblock Hybrid neuro and rule-based part of speech taggers.
\newblock In {\em Proceedings of the 18th International Conference on
  Computational Linguistics (COLING'2000)}, pages 509--515.

\bibitem[\protect\citename{Murata \bgroup et al.\egroup
  }2000]{murata_coling2000}
Masaki Murata, Kiyotaka Uchimoto, Qing Ma, and Hitoshi Isahara.
\newblock 2000.
\newblock Bunsetsu identification using category-exclusive rules.
\newblock In {\em COLING 2000}, pages 565--571.

\bibitem[\protect\citename{Murata \bgroup et al.\egroup
  }2001a]{murata_acl01_modal}
Masaki Murata, Kiyotaka Uchimoto, Qing Ma, and Hitoshi Isahara.
\newblock 2001a.
\newblock Using a support-vector machine for {Japanese}-to-{English}
  translation of tense, aspect, and modality.
\newblock {\em ACL Workshop on the Data-Driven Machine Translation}.

\bibitem[\protect\citename{Murata \bgroup et al.\egroup
  }2001b]{murata_nlc2001_wsd_eng}
Masaki Murata, Masao Utiyama, Kiyotaka Uchimoto, Qing Ma, and Hitoshi Isahara.
\newblock 2001b.
\newblock Experiments on word sense disambiguation using several
  machine-learning methods.
\newblock In {\em IEICE-WGNLC2001-2}.
\newblock (in Japanese).

\bibitem[\protect\citename{Ristad}1997]{ristad97}
Eric~Sven Ristad.
\newblock 1997.
\newblock {Maximum Entropy Modeling for Natural Language}.
\newblock ACL/EACL Tutorial Program, Madrid.

\bibitem[\protect\citename{Ristad}1998]{ristad98}
Eric~Sven Ristad.
\newblock 1998.
\newblock {Maximum Entropy Modeling Toolkit, Release 1.6 beta}.
\newblock http://www.mnemonic .com/software/memt.

\bibitem[\protect\citename{Taira and Haruno}2000]{taira_svm_eng}
Hirotoshi Taira and Masahiko Haruno.
\newblock 2000.
\newblock Feature selection in svm text categorization.
\newblock {\em Transactions of Information Processing Society of Japan},
  41(4):1113--1123.
\newblock (in Japanese).

\end{thebibliography}
\end{document}